\documentclass[runningheads]{llncs}
\usepackage{graphicx}
\usepackage{multicol}
\usepackage{caption}
\usepackage{subcaption}
\usepackage[misc]{ifsym}

\begin{document}
\title{Analyzing domain shift when using additional data for the MICCAI KiTS23 Challenge}
%
%
\author{George Stoica \Letter \inst{1,}\inst{2}, 
Mihaela Breaban  \inst{2} \and
Vlad Barbu \inst{1}}

%
\authorrunning{Stoica et al.}
%
\institute{Sentic Lab, Iasi, Romania \and
 Faculty of Computer Science, ”Alexandru Ioan Cuza” University of Iasi, Romania\\
\email{george.stoica@senticlab.com}}
\maketitle

\begin{abstract}

Using additional training data is known to improve the results, especially for medical image 3D segmentation where there is a lack
of training material and the model needs to generalize well from few
available data. However, the new data could have been acquired using
other instruments and preprocessed such its distribution is significantly
different from the original training data. Therefore, we study techniques
which ameliorate domain shift during training so that the additional
data becomes better usable for preprocessing and training together with
the original data. Our results show that \textit{transforming the additional data
using histogram matching} has better results than using \textit{simple normalization}.

\end{abstract}

\keywords{3D Segmentation \and Domain Shift \and Domain Adaptation.}

\section{Introduction}

The segmentation of renal structures (kidney, tumor, cyst) has gained interest in the recent years, starting from the KiTS19 Challenge \cite{KiTS19Challenge} and continuing with KiTS21, KiPA22\footnote{ https://kipa22.grand-challenge.org/home/} and currently with KiTS23.  
The accurate segmentation of renal tumors and renal cysts is of important clinical significance and can benefit the clinicians in preoperative surgery planning. 

Deep learning leverages on huge amount of training data for learning domain
specific knowledge which can be used for predicting on previously unseen data.
In medical image segmentation, a smaller amount of training data is available
when compared to other domains of deep learning. Therefore, using additional
training data has a greater impact on the end results.

In domain adaptation, the usual scenario entails learning from a source distribution and predicting on a different target distribution. The change in the
distribution of the training dataset and the test dataset is called domain shift.
In the case of supervised domain adaptation, labeled data from the target domain is available
\cite{wang2018deep}.

The medical image acquisition process is not uniform across different institutions and CT images may have different HU values and various amount of noise depending on the acquisition device, the acquisition time and other external factors. As a consequence, distribution shifts are easily encountered and this
affects models that perform well on validation sets but encounter different data
in practice. Creating a model which is robust to different types of distributions
requires training on enough data, coming from all the target domains.

When training under domain shift and using two datasets with different
distributions, the data should be preprocessed in order to mitigate the data mismatch error which happens due to the distribution shift. Characteristics of the
target dataset have to be incorporated into the training dataset, which could be
done either by collecting more data from the target distribution, or by artificial data synthesis. As the amount of training data from the target distribution
(KiTS23 challenge) is limited, our solution consists of transforming the additional data (taken from the KiPA22 challenge) to the target distribution. We
compare two transformations, dataset normalization, which preserves the original but different distribution, with histogram matching, which makes both the
source and target distribution the same.

\section{Methods}

Our approach consists of applying initial preprocessing to an additional dataset which was used for training. 
The aim of the preprocessing was to reduce the distribution shift between the additional data and the target domain, and will be fully-detailed in Section \ref{ch:preprocessing}.
After bridging the distributions of the original and additional data, we preprocess and normalize the whole data together and train a 3D U-Net \cite{cciccek20163d} using multiple data augmentation techniques. 
Ultimately, we evaluate our model on the validation set.

\subsection{Training and Validation Data}

Our submission uses the official KiTS23 training set, built upon the training and testing data from the KiTS19 \cite{KiTS19Data} and KiTS21 competitions.
In addition to the official KiTS23 data, our submission made use of the public KiPA22 competition training set \cite{he2021meta} \cite{he2020dense} \cite{shao2011laparoscopic} \cite{shao2012precise}.

The KiTS23 training dataset contains 489 CTs which include at least one kidney and tumor region and usually include both kidneys and optionally one or more cyst regions. 
In contrast, the KiPA22 training data contains only 70 CTs in which only the diseased kidney is selected. 
KiPA22 images have 4 segmentation targets: kidney, tumor, artery and vein. 
Unlike KiTS23, benign renal cysts are segmented as part of the kidney class for KiPA22. 
The initial preprocessing for the KiPA22 images consists of removing the artery and vein segmentation masks and keeping only the kidney and tumor class. 

We have used 342 random images from KiTS23 and 70 images from KiPA22 for training and 147 random images from KiTS23 for validation.

\subsection{Preprocessing}
\label{ch:preprocessing}

Initial exploratory data analysis illustrate the fact that images from KiPA22 have a totally different distribution than images from the KiTS23 training set on the HU scale. 

\begin{figure}[ht]
    \centering
    \begin{subfigure}[b]{0.31\textwidth}
        \centering
        \includegraphics[width=\textwidth]{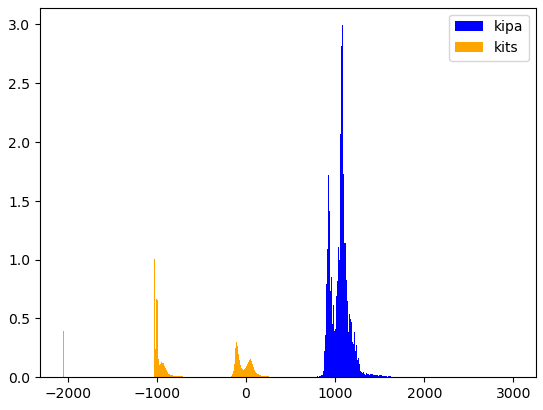}
        \caption{Both datasets before preprocessing.}
        \label{fig:histogram:initial}
    \end{subfigure}
    \begin{subfigure}[b]{0.31\textwidth}
        \centering
        \includegraphics[width=\textwidth]{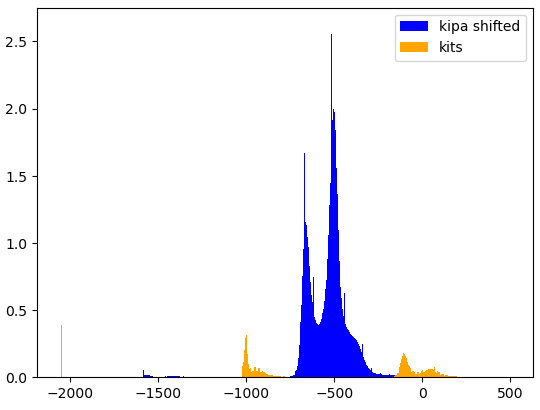}
        \caption{Shifting the mean and standard deviation.}
        \label{fig:histogram:shifted}
    \end{subfigure}
    \begin{subfigure}[b]{0.305\textwidth}
        \centering
        \includegraphics[width=\textwidth]{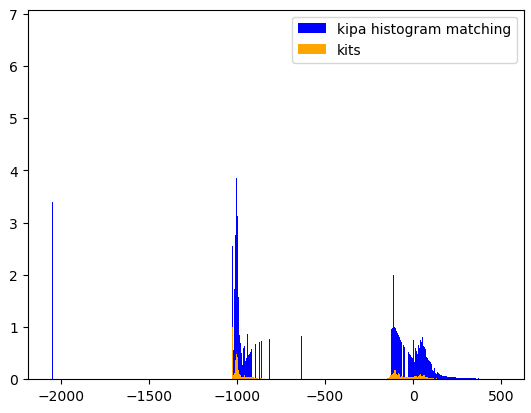}
        \caption{Applying histogram matching.}
        \label{fig:histogram:matching}
    \end{subfigure}
    
    \caption{Histograms for both datasets before and after initial preprocessing. }
    \label{fig:histogram}
\end{figure}

While the values for the KiTS23 CT images are mostly centered around -
1000 and 0 on the HU scale, KiPA22 images are situated between 800 and 1500
while also having a visible different distribution (Figure \ref{fig:histogram:initial}). 
Training under
domain shift using the original distribution for the second dataset is challenging,
therefore we have taken steps towards ameliorating the effects of distribution
shift.

To mitigate the impact of the huge distance between the values of the vertices,
the simplest solution is shifting the mean and standard deviation of KiPA images
to match those of KiTS (Figure \ref{fig:histogram:shifted}). 
Nevertheless, the distributions are visibly
different, therefore we also applied histogram matching to transform the KiPA
images to the KiTS domain (Figure \ref{fig:histogram:matching}).

We choose to transform the KiPA images because the test data will be made
of images which are expected to match the original KiTS distribution. When
training under domain shift, only data from the target distribution should be
used for validation.

For the KiPA dataset, shifting the mean maintains the original shape of the
curve, scaled by the factor which changed the standard deviation, spreading the
values evenly. Histogram matching, on the other side, is destructive and the
HU values of vertices are spread unevenly to match the KiTS distribution. To
choose the best transformation, we have created two datasets to evaluate them
separately in order to choose the more suitable one.

\begin{enumerate}
    \item \textbf{Dataset 1}: 342 images from KiTS and 70 images from KiPA whose values are shifted by changing the mean and standard deviation.
    \item \textbf{Dataset 2}: the same 342 images from KiTS and 70 images from KiPA whose values are transformed by histogram matching.
\end{enumerate}

For both datasets, the same preprocessing steps are applied, using the nnUNet framework \cite{isensee2021nnu}.
Values are clipped at the 0.5th and 99.5th percentile to remove outliers. 
Then, images are normalized to have the mean 0 and standard deviation 1 and three order-interpolation is used to resample all images into a space of $0.7636 \times 0.7636 \times 0.7636$ $mm^{3}$.


\subsection{Proposed Method}


After preprocessing each dataset, we have trained the model using the default
nnUNet configuration for training, which uses a classic 3D U-Net. Instead of
using an ensemble of 5 models trained on 5 folds of the data, we have opted to
train a single model on all the training data available.

We have used region based training, defining the 3 learning targets: Kidney
\& Tumor \& Cyst, Tumor \& Cyst and Tumor. While this approach directly optimizes the official evaluation metrics, it does not yield good results for predicting
the cyst class. We have experimented with Dice \& Focal Loss and Dice \& Cross
Entropy Loss, the latter achieving the best results.

We have trained the model using a patch size of $128 \times 128 \times 128$ and a batch
size of two, for 1000 epochs. 
We started the training using SGD and Nesterov
momentum with a learning rate of 0.01 and used a Polynomial Learning Rate
Scheduler to decrease the learning rate evenly until it reaches a value of 0.001. To
prevent overfitting, we applied multiple data augmentation techniques integrated
in nnUNet: Rotation, Scaling, Gaussian noise, Gaussian blur, Random brightness, Gamma Correction and Mirroring. We did not apply any post-processing
on the prediction.

\section{Results}




We have trained on both \textbf{Dataset 1} and \textbf{Dataset 2} and have used 147 images from KiTS23 for validation. 
The results are displayed in Table \ref{table:results:validation}. 
The official metrics used in competition are in \textit{italic}, but we also report the Dice score for kidney and cyst segmentation.

\begin{table}[ht]
\centering
\caption{Validation results for \textbf{Dataset 1} and \textbf{Dataset 2}.}
\label{table:results:validation}
\begin{tabular}{|l|l|l|l|l|l|}
\hline
Dataset & \textit{Dice kidney}\&\textit{masses} & \textit{Dice masses}  & Dice kidney & \textit{Dice tumor}  & Dice cyst \\ \hline
Dataset 1 & 95.310904         & 79.072143 & \textbf{94.101086}  & 76.891783  & 17.012944 \\ \hline
Dataset 2 & \textbf{95.453839} & \textbf{80.760511} & 94.024749  & \textbf{78.960431} & \textbf{20.766421} \\ \hline
\end{tabular}
\end{table}

Our experiments show that applying histogram equalization (\textbf{Dataset 2}) on the additional dataset improves the results for all the target metrics. 
Using simple normalization (\textbf{Dataset 1}) has better results only when calculating the Dice score for the kidney area.
However, the Dice score for tumors and cysts is worse by 2 and 3 percent. 
Using the original distribution of the KiPA dataset results in a lower Dice score for cysts also because possible cysts are labeled as the kidney class. 
Nonetheless, since the two distributions are still very different even after normalization and preprocessing, the scores are heavily impacted.

Evaluating the results on both configurations, the model does not distinguish the cyst class and many cysts are classified as tumors. 
There is a class imbalance between cysts and tumors, as cysts generally encompass a smaller area.
In our case, the low dice score for cysts is a result of many false positives. 
We presume that our learning target is the culprit, because we directly minimize the Dice and Cross Entropy loss for the whole segmentation area (kidney and masses), the lesion area (masses, both tumor and cyst) and ultimately, tumor.
As a consequence, the cyst area is indirectly learnt, therefore the accuracy is lower. 
To make the model discriminate between the two classes and reduce the false positives, we suggest changing the learning target to directly minimize the Dice and Cross Entropy loss for the cyst class.

For training and inference we have used a workstation with an RTX 3090 GPU, an AMD Ryzen Threadripper 2970WX 24-Core Processor CPU, SSD and 31GB RAM memory available. 
Training the model took around 3.4 days. 
For prediction, the average inference time was 10 minutes per case.

\section{Discussion and Conclusion}


We have explored suitable techniques for mitigating distribution shift when using additional data for the kidney tumor 3D segmentation task. 
We identify histogram matching as an initial preprocessing step of artificial data synthesis that completely transforms the additional distribution to the target domain.
Compared to simple normalization, this approach has the advantage of training only on the target distribution, which improves the results, especially for the least frequent classes, cyst and tumor.

We believe that more stable results can be achieved by training an ensemble, and the discriminative power between cysts and tumors can be enhanced by changing the training target and using techniques that deal with class imbalance.

%
%

\bibliographystyle{splncs04}
\bibliography{paper9}

\end{document}